# A hybrid feature learning approach based on convolutional kernels for ATM fault prediction using event-log data

Víctor Manuel Vargas[a], Riccardo Rosati[b,*], César Hervás-Martínez[a], Adriano Mancini[b], Luca Romeo[c], Pedro Antonio Gutiérrez[a]

[a]*Department of Computer Science and Numerical Analysis, University of Córdoba, Córdoba, Spain*
[b]*Department of Information Engineering, Marche Polytechnic University, Ancona, Italy*
[c]*Department of Economics and Law, University of Macerata, Macerata, Italy*

**Abstract**

Predictive Maintenance (PdM) methods aim to facilitate the scheduling of maintenance work before equipment failure. In this context, detecting early faults in automated teller machines (ATMs) has become increasingly important since these machines are susceptible to various types of unpredictable failures. ATMs track execution status by generating massive event-log data that collect system messages unrelated to the failure event. Predicting machine failure based on event logs poses additional challenges, mainly in extracting features that might represent sequences of events indicating impending failures. Accordingly, feature learning approaches are currently being used in PdM, where informative features are learned automatically from minimally processed sensor data. However, a gap remains to be seen on how these approaches can be exploited for deriving relevant features from event-log-based data. To fill this gap, we present a predictive model based on a convolutional kernel (MiniROCKET and HYDRA) to extract features from the original event-log data and a linear classifier to classify the sample based on the learned features. The proposed methodology is applied to a significant real-world collected dataset. Experimental results demonstrated how one of the proposed convolutional kernels (i.e. HYDRA) exhibited the best classification performance (accuracy of 0.759 and AUC of 0.693). In addition, statistical analysis revealed that the HYDRA and MiniROCKET models significantly overcome one of the established state-of-the-art approaches in time series classification (InceptionTime), and three non-temporal ML methods from the literature. The predictive model was integrated into a container-based decision support system to support operators in the timely maintenance of ATMs.

*Riccardo Rosati is the corresponding author. E-mail: r.rosati@pm.univpm.it.
Faculty of Engineering, Via Brecce Bianche 12, 60131 Ancona, Italy.

## 1. Introduction

Predictive Maintenance (PdM) has become increasingly important in recent years in the context of Industry 4.0 and intelligent manufacturing. It helps to predict when machines are going to fail so that corrective maintenance can be carried out before failure occurs. [1, 2]. In this manner, unexpected equipment downtimes are avoided, service quality is improved, and additional costs resulting from over-maintenance are also reduced. In this fault diagnosis task, model predictions of nominal system behavior are compared with data from sensors mounted on the monitored system to detect and isolate faults [3, 4]. However, collecting condition monitoring and time-series sensor data (e.g. temperature, pressure, current and power consumption, etc.) requires the machines to be equipped with sensors and proper data platform systems. It often implies huge investments to redesign sub-systems of the devices [5]. On the contrary, the machine implicitly provides event-log information, thus reflecting real-time massive state data about the machining process and potentially related to faults and errors. Event-log information produced by the machine control or Internet Technology (IT) system can be exploited as input of the monitoring system. At the same time, the output is a warning about a potential failure or an indication of the tool's Remaining Useful Life (RUL) [6]. Due to the complexity of the data and the low correlation between event-log input data and potential failure, few works [5, 6] in the state-of-the-art considered the possibility of performing failure prediction from event-log data.

In this context, detecting early faults in automated teller machines (ATMs) has become increasingly important because these machines are susceptible to various types of failures, which are complex for maintenance personnel to predict. The high complexity of the construction and installation of ATMs makes it impossible to equip the machine with specifically dedicated sensors. Indeed, the internal operating status of ATMs is recorded with a high sampling rate in the form of event logs or system messages, as well as all errors and warnings. The failure events annotation is crucial for monitoring an ATM's condition and detecting potential problems before they occur. However, the collected event-log data are not directly related to the failure event, thus requiring a complex learning procedure to model hidden discriminative patterns. When dealing with this vast amount of annotated complex data, supervised Machine learning (ML) and Deep Learning (DL) techniques seem to be the most promising approaches to capture non-linear dependencies between input and output, exploiting labeled data in the form of Multivariate Time Series (MTS) to train classification models for fault prediction [7].

Moreover, predicting machine failure based on a continuous stream of event logs poses additional challenges, mainly related to extracting features that might represent sequences of events indicating impending failures [8]. Since the performance of the final model is strictly dependent on time-series properties, the extracted features have to effectively capture different spatiotemporal patterns that are informative for the final predictive task [9]. In past literature for fault detection [10, 11], this issue was addressed by performing manual feature extraction and selection, which results in poor model generalisation capability, also due to the lack of capturing hidden patterns in the data. To overcome this limitation, feature learning approaches based on DL [12, 13, 14] are currently being used in PdM, where informative features are learned automatically from minimally processed data. However, a gap remains on how feature-learning approaches can be exploited for deriving relevant features from event-log-based data.

In this direction, recent non-deep learning methods for MTS classification, such as ROCKET [15], its evolution MiniROCKET [16] and HYDRA [17], play a crucial role for capturing relevant features from data by transforming time series using random convolutional kernels. Then, the transformed features are employed to feed a linear classifier. These methods are currently detaining the best performances for solving benchmark time series classification and regression tasks by outperforming state-of-the-art deep architectures in predictive performance and computation time [18, 19]. However, there is no evidence on how convolutional kernel approaches can be introduced to learn discriminative features for improving the generalisation performance of a real and challenging industrial task, namely the prediction of ATM's failure event using only event-log data.

This paper presents a predictive model driven by feature learning and ML algorithms for estimating the probability of ATM failure during a specified time interval. Convolutional kernel transformation is proposed to implement the feature learning procedure from the event-log data. At the same time, a ML model was introduced to provide accurate failure prediction from the learned features. Experimental results on a real and large dataset demonstrate the effectiveness of the proposed approach to other state-of-the-art methods widely employed in this context.

In particular, the main contributions of the present work can be summarized as follows:

- the application of time series classification to PdM by using event-log data for MTS classification of ATMs failure events;

- the proposal of a novel feature learning approach based on a convolutional kernel to learn discriminative features and hidden patterns from log-based time series data which are apparently unrelated to the failure event;

- the proof of concept of the proposed methodology on a large real-world challenging dataset, where time series data belong to different ATMs with an extremely not homogeneous and limited number of failure cases;

- the design of a container-based PdM platform that integrates the proposed predictive model as the main core of a decision support system.

In summary, this paper presents a novel approach based on computational intelligence to predict ATM failure events by combining advanced feature extraction, feature learning, and machine learning techniques. The rest of this paper is organised as follows: in Section 2, we review PdM and feature learning approaches. We describe the employed dataset and the proposed feature extraction procedure in Section 3, while in Section 4 we discuss the proposed approach and experimental procedure. Results and the proposed DSS architecture are presented in Section 5. Discussions are presented in Section 6. Finally in Section 7, we provide conclusions of our findings.

## 2. Related work

We review two threads of relevant prior work. The former is the body of work on data-driven methodologies for PdM and fault prediction (see Section 2.1). The latter focuses on feature learning approaches (see Section 2.2).

### 2.1. Data-driven approaches for Predictive Maintenance and fault prediction

Within the paradigm of Industry 4.0, in recent years, data-driven models have started to investigate the relation between monitoring data and machines' failure or degradation to plan PdM actions [20]. Data-driven PdM approaches were designed in different domains ranging from food processing [21], energy production [22], smart manufacturing [23], transportation industry [24] and air quality control [25]. In particular, these models can be classified into sensor-based or log-based approaches. Sensor-based approaches use time-series signals of single or multiple sensors to assess the RUL or time to failure. Sensor-based PdM typically involves monitoring individual components that are equipped with multiple sensors. Such approaches are commonly used in mechanical engineering, particularly for rotary machines or components such as bearings or gearboxes [26]. Instead, log-based approaches use historical event-log data to train an ML/DL algorithm [5]. This type of monitoring is mainly applied in IT systems like ATMs [27, 8, 28], where the aim is to monitor the machines without the possibility of installing sensors on each component. Event-log data are extracted from log messages such as system messages, alarm codes, numerical values, keywords, or machine IDs. Thus, one of the relevant aspects of log-based PdM is that many different components are monitored via the same data stream to assess the health condition of the whole system, making the final task more complex [29]. In particular, this approach presents two main challenges: i) identifying patterns in a series of discrete events, where multiple physical phenomena can trigger the same event log, and ii) extracting relevant features, considering that only relatively small portions of the datasets are strictly related to PdM issues.

Early ML approaches for event-log MTS classification are based on handcrafted features, typically with sliding window statistics for managing the temporal constraint and standard ML algorithms [8]. However, manual feature engineering may cause the loss of detailed and useful information, not allowing the discovery of possible hidden data patterns [30, 31]. Our approach can be categorized as a hybrid feature extraction method [32], as it integrates a feature learning procedure (convolutional kernel) with extracted features from event-log data. This proposal offers the advantage of creating a more representative input space, which a simple machine learning model can use to provide prompt and precise predictions of failure events. In recent years, DL models have been increasingly used in a supervised manner to develop high-performance RUL prediction and early fault detection algorithms, representing an effective technique for automatically performing feature extraction from raw time-series data through multiple non-linear transformations [33]. DL methods such as Convolutional Neural Network (CNN) or Recurrent Neural Network (RNN) are used in the state-of-the-art to model the temporal dimension mainly for regression tasks and time series forecasting [34]. RNN models have been widely applied to RUL prediction regression task since they can learn the temporal relationships with the help of the recurrent architecture [35, 36]. However, sequential approaches suffer from the vanishing gradient problem as the sequence length increases [37], demonstrating the lack of modeling long-term temporal dependency, which is a crucial point in maintenance cases. Moreover, unlike our approach, the sequential learning model requires the knowledge of time-series predictors that disclose long and short-term relationships between the failure event and the degradation process. For this reason, recent frameworks have proposed using Transformers architectures based on the attention mechanism to model the relationships between pre-extracted low-level features and RUL regression [38]. Although the attention-based architecture integrated into the transformers has been proven to learn discriminative and interpretable degradation patterns, several challenges remain in the practical use of these models in the actual industry scenario. Often, standard optimisation techniques for tuning the most sensitive hyperparameters (e.g. embedding dimension, number of heads, and number of layers) lead to sub-optimal performance and a high computation effort [39]. Hence, the application of Transformer-based approaches is limited for predicting failure events using event-log data, which can be of both high dimension and long-length.

*2.2. Feature learning approaches for time series classification*

Moving towards feature learning approaches for MTS classification, DL methods based on CNN and Autoencoders were primarily used for critical feature extraction and fault detection in PdM tasks [40, 41]. When compared to traditional machine learning models, such as Support Vector Machine (SVM), k-Nearest Neighbor (kNN), and Decision Tree (DT)-based methods, CNN architectures can attain superior fault detection outcomes by learning spatiotemporal patterns, without necessitating prior handcrafted feature extraction. In particular, InceptionTime [42] is an ensemble of CNNs based on the Inception architecture and is one of the most accurate CNN models for time series classification in different domains [19]. However, the InceptionTime has not already been used for the PdM classification task based on event logs from MTS data.

Currently, non-deep learning methods such as HIVE-COTE [43], TS-CHIEF [44], and ROCKET [15] leverage convolutional kernels to learn complex discriminative patterns on the input data. These methodologies constitute the state-of-the-art MTS classification according to different evaluations on public benchmarks [45]. Recent literature demonstrates the simplicity and the higher convergence rate of convolutional kernels than the InceptionTime model [15]. ROCKET, which is the best ranking method on average, involves training a linear classifier on top of features extracted by a flat collection of numerous and various random convolutional kernels. Later variants of this approach aim to improve processing time on large datasets further while achieving the same performance. MiniROCKET [16] is a variant of ROCKET which is up to 75 times faster than ROCKET while offering essentially the same accuracy in the benchmark datasets. HYDRA [17] is a fast and accurate dictionary method combined with convolutional kernels.

These feature learning approaches based on convolutional kernels have recently been proposed for solving different tasks [46, 47, 48, 49], but never introduced to deal with complex PdM problems such as fault detection prediction using event-log data. Introducing these convolutional kernel techniques is helpful in learning features that can spot failures from event-log data. Moreover, extracting powerful predictors may lead to completely decoupling the feature extraction phase from the prediction phase, which can also be implemented using a naive linear ML model. To fill this gap, the paper introduces fast convolutional kernel methods, i.e. MiniROCKET and HYDRA, that can learn discriminative features from the collected ATMs event-log data. Our approach addresses a critical gap in the literature by using event-log data to solve a PdM task of fault detection prediction in an unexplored domain of banking automation. Thus, our methodology presents a significant novelty that can be applied to real-world engineering problems in the challenging field of Industry 4.0.

## 3. Materials

In this section, the problem presented by the ATM manufacturing company and the data they have collected from the usage of their machines are described. Additionally, the process of creating a dataset from this information is outlined. Specifically, in Section 3.1, the data provided by the company and the problem they aim to solve are detailed. Section 3.2 describes the steps taken to generate a dataset that can be used to train the proposed approach to address the problem. The idea for the overall project originated from the demand of Sigma S.p.A., an actual ATM company. Data regarding customers and instruments are anonymous, and an agreement between Sigma S.p.A. and the University regulates their use, detention and conservation. [1]

### 3.1. Raw data and real-world problem

ATMs generate a large volume of logs that can contain all the information about the operation of the devices. These logs are often written in verbose mode to allow for the analysis of any abnormal behavior. However, they are not typically structured according to strict rules. Given that the company that manufactures the ATMs is also responsible for their maintenance and they have a large number of machines, predicting failures can be challenging because they are often the result of each machine's usage and specific circumstances. For this reason, some machines experience failures more frequently than others. In an effort to use ML to predict the failure events of these machines, the company has saved the raw logs of 68 machines that were generated for almost two years (2020-2022) along with the annotations that the operators did when they performed a maintenance task on any of the machines. In this way, two different files were provided by the company:

---

[1] Authors will publish the anonymised datasets in a public repository.

1. First, they gathered the "states file" of the machines, aggregating the event logs of all the machines considered for the problem described. Thus, each line of this file represents an operation performed in a specific machine along with the result of the operation. In this way, each row includes an identifier for the machine, the date and time of the operation, the type of operation, and the response obtained from the machine (successful operation or error). These logs are automatically generated by the machines and compiled into a single file by the company. An example of the information contained in the states file can be seen in Table 1. Note that not all the errors found in this file are associated with a failure in the device. Most errors are associated with customer misuse and do not represent a problem with the machine. Some failures can be identified by only checking the log file, given that the machine stops working and the error is repeated multiple times in the log file. However, there are other types of failures where the machines keep working, and the problem appears periodically. In these cases, the problem is hard to identify. For that reason, the company also provided the annotations file, which points out those failures that required the operator to fix them. It is worth noting that the states file is merely an event log and does not contain information from any machine sensor.

2. Secondly, the company has provided a file that lists the maintenance tasks performed on each machine. This file includes the machine's identifier, the maintenance date, and the type of maintenance performed. This "annotations file" is manually created by the machine's operator performing maintenance. Not all of the issues listed in this file are related to failures; other types of issues, such as preventive maintenance tasks, are also included. Thus, the failure events annotated in this file are classified into five categories, apart from the preventive maintenance: the existence of a foreign body inside the ATM, generic failure of any of the components of the machine, jam, replacement of any of the devices of the machine, and wrong usage. The most important limitation of the annotations provided in this file is that it contains the maintenance task date but does not include the specific time when it was performed. Moreover, inferring the failure date from the maintenance task's date is not trivial, given that the ATMs can start experiencing issues several days before the operator is called. For reference, some rows of this annotations file can be seen in Table 2.

*3.2. Features extraction*

Concerning the feature extraction technique, our method analyses and extracts representative information for semi-structured data originated from ATM banking. Unlike other state-of-the-art works, the raw data is not used as free text. The data is formed by states files which directly represent the type of operation (command), and the response obtained from the machine (response) (see Table 1). The feature extraction step ensures to compute a time series that represents the number of occurrences of the associated command and response since the last failure event, i.e. in this cycle of failure. Therefore, the time series value is accumulated until the next cycle begins.

| Machine | Date | Time | CmdType | RespErr |
|---|---|---|---|---|
| 0487 | 2021-04-03 | 07:49:55 | PrepareWithdrawal | ResponseOk - 0x0000 |
| 0990 | 2021-12-24 | 14:44:38 | CountNote | NVNoteSerialNumberError - 0x3510 |
| 1395 | 2021-03-11 | 20:31:38 | Initialize | NoteRemainInCashSlot - 0x0611 |
| 1549 | 2021-01-19 | 07:07:09 | CloseShutter | ResponseOk - 0x0000 |
| ... | ... | ... | ... | ... |

Table 1: States file example. The CmdType column shows different commands that the ATM can run, while the RespErr column shows the response obtained from the machine. The 0x0000 code is associated with a successful operation, while any other code represents an error.

| Machine | Date | Type of maintenance task | | | | | |
|---|---|---|---|---|---|---|---|
| 487 | 2020-06-16 | 0 | 1 | 0 | 0 | 0 | 0 |
| 739 | 2021-05-06 | 0 | 0 | 0 | 0 | 1 | 0 |
| 1549 | 2021-11-23 | 0 | 0 | 0 | 0 | 0 | 1 |
| ... | ... | ... | ... | ... | ... | ... | ... |

Table 2: Annotations file example. The type of maintenance task column represents with 0 or 1 the different types of maintenance tasks described, in the same order: foreign body, generic failure, jam, preventive maintenance, replacement of any part and bad usage.

The raw data described in the previous section was used to create a dataset for predicting a failure event for any given ATM using the proposed feature learning approach and machine learning model. The states file was divided into parts for each machine, and a list of all the days the machine was operating was created. Each day was divided into 10-minute intervals, and for each interval, the number of correct and error responses obtained for each type of command was counted. The choice of 10-minute intervals was made to ensure that at least one ATM transaction by the customer was captured within each interval, according to the company's experience. The computed features within these intervals may be potentially correlated with other operations carried out at different time points and can predict failure events. Expanding the granularity beyond this limit would raise the dimensionality of the data with no significant improvement in the dataset's information, as numerous intervals would not contain any transactions. Conversely, using wider intervals could result in the loss of crucial information and potential correlations between operations performed at various time instants.

This procedure resulted in 38 values for each time interval and machine: 2 types of responses (successful operation or error) for each of the 19 types of commands. The time intervals of the same day and machine were grouped to create a MTS with 38 dimensions, each with a length of 144 time intervals per day.

The label for each sample was obtained from the annotations file. The file was filtered to remove annotations related to preventive maintenance tasks, as these tasks do not indicate a failure in the machine. The label for each sample was calculated by finding the number of days between the sample's date and the date of the next failure for that specific machine. However, the labels obtained using this annotations file are prone to inaccuracies derived from two main sources:

- Lack of time information in the annotations: it is not possible to know the exact time of the failure, only the date. Sometimes, the date may also be inaccurate, as the failure can occur several days before the operator is called for maintenance. Therefore, labeling samples that are very close to a failure event may be inaccurate.

- Human error: the annotations are manually written by the operator, which can result in mistakes when specifying some data.

For these reasons, accurately predicting the exact time (continuous value) before the next failure event occurs is not possible, and, therefore, we are assigning a binary label to each sample by grouping the remaining days before the failure event in two groups: the first contains those samples of machines that will experience a failure in less than one week and the other contains the rest of the samples. Consequently, the task can be approached as a binary classification problem using machine learning techniques. Moreover, the company's requirements influenced the decision to create a binary label rather than using the precise number of days. The company is primarily interested in predicting imminent failures instead of determining the exact remaining time before a distant failure event. Additionally, the time interval of one week was chosen to achieve a consistent prediction, which can aid reactive maintenance as per the company's demand. By using one-week intervals, an alert can be generated in close proximity to the failure (e.g., one day before the failure) or by anticipating the failure consistently, thereby enabling the operator to intervene for proper maintenance. These intervals also represent the best trade-off for obtaining a representative number of positive samples. Moreover, empirical results from previous studies [28] confirmed the drawbacks of designing a multi-class approach in this context. Indeed, if we only consider the fault class as occurring on day 1 of RUL, this implies a small number of failures. From a practical perspective, this leaves little time to model and prevent failures. On the other hand, if we consider the fault class several days before day zero of RUL, there is a high risk of training the ML model on data that are not sufficiently discriminative of failures. In other words, the data may be too far removed from actual failures.

To summarise, the constructed dataset contains 29, 386 samples representing one day of one machine. In this way, each sample is an MTS with 144 points, associated with all the 10-minute time windows of that day and 38 dimensions. Each dimension is related to a command type and response type, where the value of each point in the time series represents the number of occurrences of the associated command and response since the last failure event., i.e. in this cycle of failure. Therefore, the time series value is accumulated until the next cycle begins. The label of each sample, obtained from the annotations file, determines whether a failure event will happen in less than one week. Additionally, some extra information is included for each sample, such as the cycle of failure and the machine identifier, which can be used to create data partitions without mixing different machines. The process of creating this dataset is shown in Figure 1.

The dataset described above and the binary labels assigned to each sample represent a novel binary classification problem that can be addressed using ML and DL techniques. The goal of the problem is to predict whether a failure will occur within the next 7 days, using only the information from the current day (accumulated since the last error), which is based on an event-log. This problem is challenging because failures sometimes show different patterns of errors. Sometimes an error is represented by a block of consecutive errors in the log file, while other failures are characterised by isolated errors that repeat at a certain frequency. Additionally, the inaccuracies in the annotations file used to label the samples add further complexity to the problem. Finally, another point to keep in mind when addressing this problem is the limited number of failure cases with respect to the number of correct cases, giving birth to quite an imbalanced problem where only 3064 samples belong to the positive class. In contrast, the negative class contains 26, 322 samples.

## 4. Methods

In this section, the feature learning and predictive models used (see Section 4.1), as well as the experimental comparisons (see Section 4.2) and experimental procedure (see Section 4.3) followed, are described.

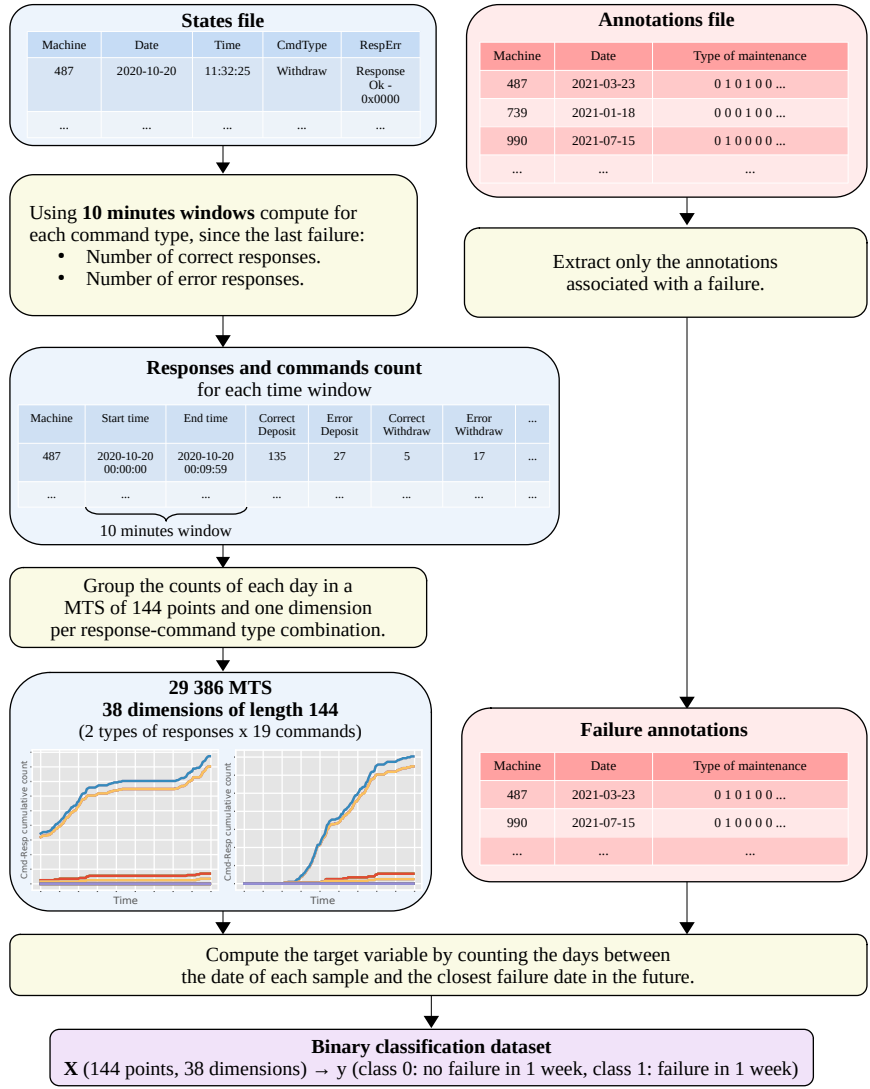

Figure 1: Dataset creation process: each of 29,386 sample represents an MTS with 144 points, associated with all the 10-minute time windows of that day and 38 dimensions. Each dimension is associated with a command type and response type. The value of each point in the time series represents the number of occurrences of the associated command and response since the last failure event, i.e. in this cycle of failure. Therefore, the time series value is accumulated until the next cycle begins. The label of each sample, obtained from the annotations file, determines whether a failure event will happen in less than one week.

*4.1. Feature learning and Predictive models*

Note that no information related to the failure event is lost during the construction of the time series dataset. Although a feature extraction step is applied, it is important to ensure that the resulting input space is fully discriminative of the failure event. In this way, we propose to use feature learning approaches based on convolutional kernels to address the problem described in Section 3. Three different predictive models that fall into this category and are well-suited for MTS are employed and described below:

1. ROCKET [15] transforms input time series using many random convolutional kernels (by default, 10, 000) and uses the transformed features to train a linear classifier. The lengths of the kernels are randomly selected from $\{7, 9, 11\}$, the weights are obtained from a normal distribution, and the biases are drawn from a uniform distribution $[-1, 1]$. The dilation and the padding are also randomly selected. It applies global max pooling and "proportion of positive values" (PPV) pooling to the convolution output. The resulting features are used to train a linear classifier.

2. MiniROCKET [16] was proposed as an improvement to the ROCKET method [15]. MiniROCKET makes several changes to ROCKET, intending to remove randomness and significantly speed up the transformation. According to this, MiniROCKET uses a small, fixed set of kernels and is almost entirely deterministic. The kernel parameter values were validated in [16] and motivated by theoretical and empirical proof. In particular, MiniROCKET uses kernels of length 9, with weights restricted to two values, building on the observation made in [15] that weights drawn from $\{-1, 0, 1\}$ produce similar accuracy to weights drawn from $(0, 1)$. There are $2^9 = 512$ possible two-valued kernels of length 9. MiniROCKET employs a subset of 84 of these kernels as a trade-off between performance and computation effort [15, 16]. Then, it uses bias values drawn from the convolution output, a fixed set of dilation values, and only produces PPV features. Lastly, a linear classifier is trained using the transformed features. These changes allow removing almost all randomness from ROCKET, reducing the time needed for the transformation process.

3. HYDRA [17], a dictionary method that uses convolutional kernels and incorporates aspects of ROCKET and conventional dictionary methods for time series classification. It involves transforming the input time series using a set of random convolutional kernels arranged into *g* groups with *k* kernels per group. At each time point, the number of kernels representing the closest match with the input time series is counted for each group. Unlike typical dictionary methods, HYDRA uses random patterns represented by random convolutional kernels. Thus, HYDRA inherits major kernel characteristics from Rocket and MiniROCKET, such as the kernel length of 9, a weight drawn from normal distribution N(0, 1) and a fixed exponential dilation. Therefore, unlike MiniROCKET, the kernels are organised into groups and HYDRA treats each kernel as a pattern in a dictionary by treating each group as a dictionary. Consequently, HYDRA uses two different forms of counting: wherever the convolution output is positive hard counting implies incrementing the count for every kernel, and soft counting means accumulating the output values for each kernel. The counts are then used to train a linear model.

In our case, we used the Ridge classifier [50] as the linear classifier, as it offered the best balance between simplicity, performance, and computational effort. This finding is consistent with previous work [15, 17].

*4.2. Experimental comparisons*

InceptionTime [42] was employed as a comparison, as the most accurate state-of-the-art model for time series classification in different domains [51, 52]. InceptionTime is a DL model based on the Inception-v4 model [53]. The InceptionTime architecture comprises an ensemble of five Inception networks, with each prediction given an even weight. Each Inception network contains two different residual blocks. Each block is comprised of three Inception modules. Each residual block's input is transferred using a linear connection to be added to the next block's input, thus mitigating the vanishing gradient problem by allowing a direct flow of the gradient [54]. The residual blocks are followed by a global average pooling that averages the output MTS over the whole time dimension. At last, the Inception network contains a traditional fully-connected softmax layer with several units equal to the number of classes in the problem.

In addition, three non-temporal ML approaches are considered for comparison with the approaches proposed in this work. To use the MTS input to train those ML methodologies, the MTS has been transformed into tabular data following the same approach employed in [5]. In this way, the resulting dataset contains $N \cdot L$ samples, each with $D$ features, where $N$ is the number of samples ($N$ = 29386), $L$ is the length of the time series ($L$ = 144), and $D$ is the number of dimensions in the dataset used for the temporal approaches ($D$ = 38). The features are accumulated in a 10-minutes time window. Then, Random Forest (RF) [55], Ridge, and XGBoost (XGB) [56] classifiers are considered for comparison with non-temporal approaches to determine whether constructing a time series dataset provides a significant benefit in comparison to working with non-temporal data.

*4.3. Experimental procedure*

The evaluation of our methodology in the described dataset represents appropriate steps for the applicability of the proposed approach in a real-use case scenario.[2] To test the performance of the methods mentioned above in the proposed problem, a 5-fold cross-validation scheme was used to create train and test partitions over machines. Moreover, the 5-fold was repeated six times with different seeds to achieve 30 executions. All partitions are stratified according to the class labels and grouped by machine identifiers so that the data from a machine is only used for training, validation, or testing.

The algorithms under consideration have parameters that must be adjusted based on the training data. In this case, for the HYDRA method, the number of groups, kernels per group, and the $\alpha$ used for the Ridge classifier were adjusted using a nested 5-fold cross-validation strategy and a grid search approach performed over the training set. Similarly, for the MiniROCKET method, the number of features and the $\alpha$ for the Ridge classifier were also fitted using the same strategy. In contrast, the maximum number of dilations was left at its default value, which the authors of the original work suggested. Finally, for the InceptionTime model, the values suggested in the original work were used for the number of filters and the learning rate. The number of training epochs was set to 100, sufficient to achieve a model that fully converges to the best weights. Overfitting due to the number of epochs is not an issue, as the best model based on the validation loss is always selected. To do that, 20% of the training data taken in a stratified way is used to create the validation set. Table 3 shows the hyperparameters that were considered for each of the four time series methods evaluated in this study. The parameters for the non-temporal approaches can be found in the accompanying code.

To carry out a more exhaustive analysis of the performance of the different methods, several metrics associated with binary classification problems have been considered:

1. Accuracy. Determines the ratio of samples that have been classified in the correct class. Formally, this metric can be defined as follows:

$$\text{Accuracy} = \frac{1}{N} \sum_{i=1}^{N} \mathbb{1}\{y_i = \hat{y}_i\}, \tag{1}$$

---

[2]Authors will publish the code and the anonymised datasets to run the experiments of our approach, modify the setting and reproduce the results in a public repository.

| Method | Hyperparameters | Values |
| --- | --- | --- |
| HYDRA | Number of groups ($g$) | $\{4, 8, 16, 32\}$ |
| | Kernels per group ($k$) | $\{2, 4, 8\}$ |
| | Ridge $\alpha$ | $[10^{-3}, 10^3]$ |
| MiniROCKET | Number of features | $\{250, 500, 1000, 2000, 4000\}$ |
| | Ridge $\alpha$ | $[10^{-3}, 10^3]$ |
| ROCKET | Number of features | $\{250, 500, 1000, 2000, 4000\}$ |
| | Ridge $\alpha$ | $[10^{-3}, 10^3]$ |
| InceptionTime | Number of filters | 32 |
| | Learning rate | $10^{-3}$ |
| | Number of epochs | 100 |

Table 3: Employed Hyperparameters for all temporal models. Our proposed approaches (HYDRA, MiniROCKET and ROCKET) and InceptionTime model.

where $N$ is the number of samples, $\mathbb{1}\{y_i = \hat{y}_i\}$ is 1 when the condition is met and 0 otherwise, $y_i$ is the target label and $\hat{y}_i$ is the predicted one.

2. Balanced accuracy, which is similar to the accuracy but is defined to account for the imbalance of the problem. It is defined the average recall obtained across all the classes:

$$\text{Bal-Accuracy} = \frac{1}{J} \sum_{j=1}^{N} \text{recall}_j, \qquad (2)$$

where $J$ is the number of classes (i.e. 2 in our problem), $\text{recall}_j = \frac{O_{jj}}{O_{j\bullet}}$, $O_{jj}$ are the elements of the main diagonal of the confusion matrix, and $O_{j\bullet}$ represents the sum of the $j$-th row of the confusion matrix.

3. F1 score, which is defined as follows:

$$\text{F1} = \frac{1}{J} \sum_{j=1}^{J} 2 \cdot \frac{\text{precision}_j \cdot \text{recall}_j}{\text{precision}_j + \text{recall}_j}, \qquad (3)$$

where $\text{precision}_j = \frac{O_{jj}}{O_{\bullet j}}$, and $O_{\bullet j}$ is the sum of the $j$-th column of the confusion matrix.

4. AUC is the area under the ROC curve, which is usually used to measure the performance in binary classification problems.

5. MS or Minimum Sensitivity [57], which is computed as the minimum value

obtained when computing the sensitivity for each class independently:

$$\text{MS} = \min \left\{ \text{recall}_j; j = 1, ..., J \right\}. \tag{4}$$

In addition to the described metrics, the time required to fit each model with a given set of hyperparameters and considering only one cross-validation fold is taken into account.

## 5. Results

This section shows the results of the experiments described in Section 4.3. Given that each method has been run 30 times, with six different seeds and five folds for each seed, Table 4 shows the average value obtained for each of the metrics considered. Also, the standard deviation is shown as a sub-index along with every mean value.

| Method | Accuracy ↑ | Bal-Accuracy ↑ | F1 ↑ | AUC ↑ | MS ↑ | Time (s) ↓ |
|---|---|---|---|---|---|---|
| \multicolumn{7}{c}{Temporal approaches} | | | | | | |
| HYDRA + Ridge | $0.7590_{0.048}$ | $\mathbf{0.6930_{0.033}}$ | $\mathbf{0.3486_{0.048}}$ | $\mathbf{0.6930_{0.033}}$ | $\mathbf{0.6075_{0.055}}$ | $6.5_{2.9}$ |
| MiniROCKET + Ridge | $0.7286_{0.042}$ | $0.6639_{0.024}$ | $0.3113_{0.046}$ | $0.6639_{0.024}$ | $\mathit{0.5814_{0.050}}$ | $23.3_{7.3}$ |
| ROCKET + Ridge | $0.7731_{0.024}$ | $\mathit{0.6666_{0.016}}$ | $\mathit{0.3275_{0.033}}$ | $\mathit{0.6666_{0.016}}$ | $0.5320_{0.039}$ | $99.2_{27.3}$ |
| InceptionTime | $0.7108_{0.060}$ | $0.5388_{0.041}$ | $0.1869_{0.039}$ | $0.5388_{0.041}$ | $0.3201_{0.088}$ | $227.7_{48.2}$ |
| \multicolumn{7}{c}{Non-temporal approaches} | | | | | | |
| RandomForest | $\mathit{0.8346_{0.048}}$ | $0.5197_{0.027}$ | $0.1315_{0.049}$ | $0.5197_{0.027}$ | $0.1205_{0.050}$ | $337.4_{69.7}$ |
| Ridge | $0.6661_{0.060}$ | $0.5965_{0.038}$ | $0.2430_{0.041}$ | $0.5965_{0.038}$ | $0.5029_{0.065}$ | $\mathbf{2.9_{0.2}}$ |
| XGBoost | $\mathbf{0.8411_{0.040}}$ | $0.5280_{0.023}$ | $0.1428_{0.050}$ | $0.5280_{0.023}$ | $0.1316_{0.063}$ | $62.8_{12.3}$ |

Table 4: Average test results of the 30 executions (5-fold repeated with six seeds) for each method. The sub-indices represent the standard deviation of each method and metric. The result of the best method for each metric is highlighted in bold font and the second best in italics.

The HYDRA model combined with the Ridge classifier achieved the best results in terms of Bal-Accuracy, F1, AUC, and MS metrics. XGB obtained the best average accuracy (0.8411). However, this result is strongly influenced by the imbalanced configuration of the dataset. As a result, the model classifies most of the samples in the majority class, leading to poor performance in other metrics. This issue is evident in the MS metric, where the XGB model achieved a mean value of 0.1316. A similar problem can be observed with the RF classifier, which also achieves high accuracy. Still, at the cost of greatly reducing AUC and MS. It is worth noting that the performance of these methods does not meet the company's requirements, as their sensitivity in detecting future failures is found to be insufficient, rendering the resulting model useless for practical purposes. The AUC of HYDRA + Ridge is a 3%, 4% and 15% higher to those of MiniROCKET + Ridge, ROCKET + Ridge and InceptionTime, respectively. Regarding computational time, the HYDRA classifier is the fastest among the temporal approaches, followed by MiniROCKET, and ranks second overall (Ridge is the fastest). Both HYDRA and Ridge are significantly faster than the InceptionTime model. The high standard deviation values for computational time are primarily due to the non-uniformity of the GPUs used to run the experiments. However, all methods were executed on the same set of machines, ensuring that the computational times of all methods, considering all executions, are comparable.

## 5.1. Statistical analysis

A statistical analysis has been performed for each of the metrics considered to obtain more robust conclusions concerning the results presented in the previous section. Before performing any statistical test, as a preliminary analysis, the boxplot of the AUC metric is shown in Figure 2, given that said metric is one of the most representative metrics for binary classification problems.

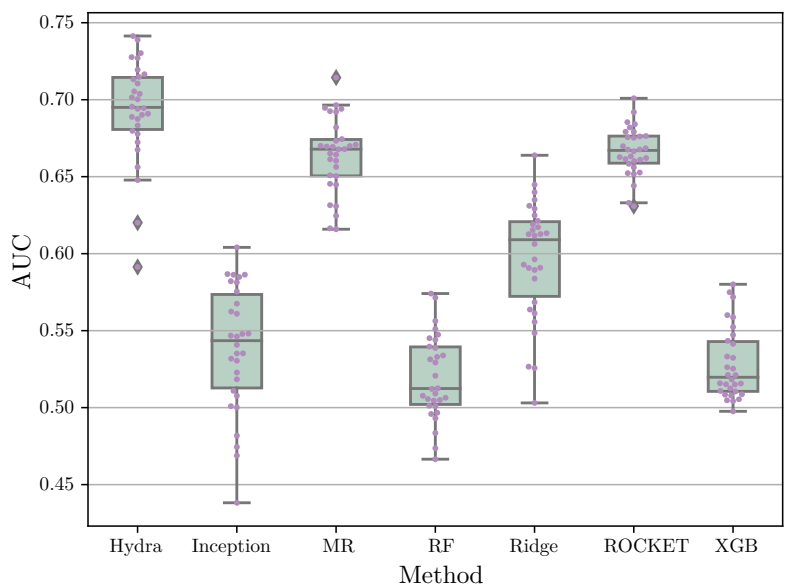

Figure 2: Boxplot for the AUC of all the temporal and non-temporal methods. Experiments were conducted by implementing six times a 5-fold CV scheme. All partitions are stratified according to the class labels and grouped by machine identifiers so that the data from a machine is only used for training, validation, or testing. All the methods were executed on the same set of machines

Given that the data partitions produced by using the 5-fold strategy are not independent, and the results of specific metrics are not normally distributed, a non-parametric statistical analysis was conducted. To validate this decision, we conducted the Anderson-Darling test [58] to check if the results were normally distributed. Our findings, with an alpha level of 0.05, show that the AUC values do not follow a normal distribution.

Thus, a Wilcoxon signed-rank test [59] was performed to compare each pair of methods considering all the metrics. Given that 7 different methods are compared, the total number of comparisons is 21. Therefore, the corrected significance (Bonferroni correction [60]) level for the Wilcoxon test is $\alpha^* = 0.05/21 = 0.0024$. The results of this test for the AUC metric can be observed in Table 5.

| Method | HYDRA | IT | MR | RF | Ridge | ROCKET | XGB |
| --- | --- | --- | --- | --- | --- | --- | --- |
| HYDRA | 1.000 | **< 0.001** | **0.001** | **< 0.001** | **< 0.001** | **0.001** | **< 0.001** |
| IT |  | 1.000 | **< 0.001** | 0.019 | **< 0.001** | **< 0.001** | 0.082 |
| MR |  |  | 1.000 | **< 0.001** | **< 0.001** | 0.926 | **< 0.001** |
| RF |  |  |  | 1.000 | **< 0.001** | **< 0.001** | 0.090 |
| Ridge |  |  |  |  | 1.000 | **< 0.001** | **< 0.001** |
| ROCKET |  |  |  |  |  | 1.000 | **< 0.001** |
| XGB |  |  |  |  |  |  | 1.000 |

Table 5: Results of the Wilcoxon signed-rank test for the AUC metric. The $p$-values of the methods which are statistically different are marked with bold font. IT: InceptionTime, MR: MiniROCKET, RF: RandomForest, XG: XGBoost.

The results of the Wilcoxon test for the AUC metric indicate that, in general, the performance of all the methods differs significantly. However, some methods are statistically similar when evaluating this metric. Thus, the performance for MiniROCKET and ROCKET is similar, as described in [16]. Also, the results for InceptionTime and RandomForest ($p$-value = 0.019), InceptionTime and XGBoost ($p$-value = 0.082), and RandomForest and XGBoost ($p$-value = 0.090) are similar. The statistical tests performed for the rest of the metrics are available in the Appendix A.

From the results and the statistical analysis described in this section, we can conclude that the HYDRA method obtained the best overall results, which are significantly higher than results obtained with the other methods. The obtained outcome can be attributed to Hydra's potential to serve as a straightforward, rapid, and precise approach for learning features, incorporating elements of both the ROCKET and traditional dictionary methods. This benefit could be advantageous in identifying concealed patterns in the event-log data that may exhibit linear correlations with the failure event. In all cases, considering Balanced accuracy, F1, AUC and MS metrics, the proposed HYDRA and MiniROCKET models always perform significantly better than the state-of-the-art InceptionTime model for solving the proposed PdM task.

*5.2. Decision Support System architecture*

A decision support system (DSS) for ATMs fault prediction was designed to integrate the predictive model described previously. The architectural diagram of the DSS platform is shown in Figure 3. The strategy of container-based deployment has already been established for managing the deployment of ML models in the Industry 4.0 domain. As the company already used this platform to manage data storage and other AI tools, we chose to implement the DSS within the existing infrastructure, which is shown in Figure 3. Container-based deployment technology ensured that individual components and the platform could easily be configured and deployed on any hardware (HW) infrastructure. Containers allow for quickly managing each application's scalability and resources and checking the integrity status. The functional macro-modules of the platform can be summarised as follows:

- Data Ingestion: it was realised by placing a Load Balancing layer via Nginx as an external entry point to the container. As communication protocols, HTTPS and MQTT with the classic publish-subscribe model are used, with VerneMQ as the MQTT server;

- Data Persistence: for the historicisation of data from the periphery, Cassandra was chosen as the NoSQL database because of its capability to manage large datasets and the availability of connectors for Kafka and Spark, while PostgreSQL was used for the structured part and historicisation of ML results;

- Data Processing: a data processing pipeline to ensure high generality and scalability of data management logic was defined. For this reason, the pipeline was based on Python and its data-processing libraries. As regards the ML model, the platform allows model training and deployment or just real-time inference. For deploying models, Spark proved to be the best candidate to implement the pipeline in production;

- Data Visualization: the final user can view model results and perform real-time monitoring of devices. At a higher level, a GUI was also built to manage the multi-user configuration and the interaction with data storage modules. Open-source tools have been preferred to create dashboards that are not only visually appealing but also fully customizable and controllable.

From an application perspective, the infrastructure modules and the related GUI allow access to all the high-level functions of the proposed PdM DSS, such as system configuration, monitoring of device status variables, control/programming of parameters, creation of custom control panels and predictive analysis reports based on event recognition and reporting (i.e. alert or failure). Additionally, the adaptation of the DSS to the company's existing infrastructure guarantees seamless integration.

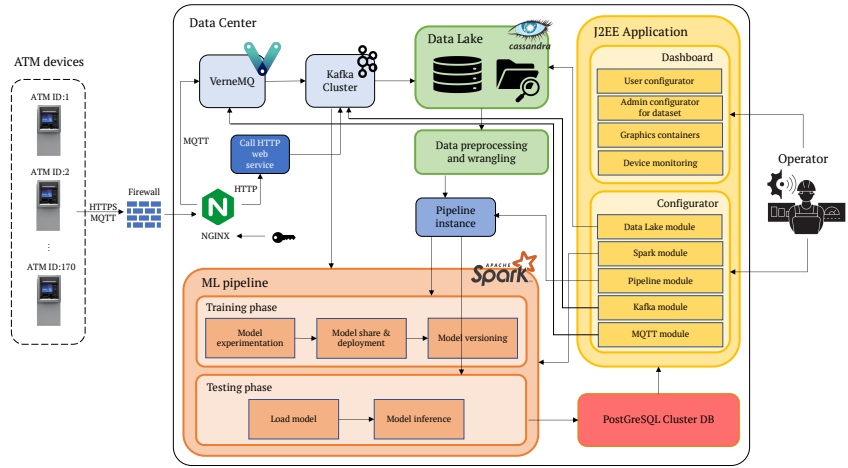

Figure 3: Architectural diagram of the PdM platform for ATMs. Data Ingestion was realized by placing a Load Balancing layer via Nginx as an external entry point to the container. Cassandra was chosen as the NoSQL database for the historicisation of data (data persistence) from the periphery. Spark proved to be the best candidate for deploying models to implement the pipeline in production. Open-source tools were chosen to create dashboards that are fully customisable and controllable.

As mentioned above company already used the described platform for managing data storage and other AI tools. What represents a novelty in this context is integrating our approach into a DSS and, more specifically, into a data processing module (see Section 5.2), which allows us to operate directly with other modules (data ingestion, data persistence and data visualization). The following characteristics also confirm the general purpose of the platform in terms of scalability and interoperability criteria:

- the DSS provides a timely predictive outcome that is disclosed through a dashboard that can support the human operator during maintenance procedure;
- the platform can be easily configured and deployed on any HW infrastructure;
- the predictive model can be triggered once new data are collected;
- the predictive model can be easily updated once new annotated data are collected.

## 6. Discussions

Table 6 illustrates the impact of our proposed approach and the challenges that were overcome through our methodology. Our hybrid feature learning approach enables the identification of hidden patterns in the event-log data that are linearly correlated with failure events, thus allowing for timely PdM. By considering container infrastructure and the described modules, we can ensure generality, flexibility, modularity, and maintainability of the centralized platform, as summarised in Table 6).

| Challenges | Methodologies | Impacts |
| --- | --- | --- |
| No discriminatory information for event-log data | Hybrid feature learning approach (feature extraction and feature learning based on convolutional kernel) | HYDRA, MiniROCKET and ROCKET capture hidden patterns in the event-log data that can be linearly correlated with the failure event. This point is also confirmed by the statistical gain of HYDRA+Ridge with respect to other competitors. |
| Support the human operator during maintenance tasks | Development of a container-based platform that integrates the DSS into data processing and data visualization modules | DSS provides a timely predictive outcome that is disclosed through a dashboard and the platform can be easily configured and deployed on any HW infrastructure |
| Need to have a timely and accurate response from the predictive model and a continuous update of the model | Development of data ingestion and data persistence modules | DSS can be triggered once new data are collected and the predictive model can be easily updated once new annotated data are collected. |

Table 6: Impact of the proposed approach in terms of challenges faced and methodologies used to solve them.

## 7. Conclusions

In conclusion, this study presents a novel predictive methodology for learning in a complex real-world dataset created from ATMs' event-log data. The proposed models are based on convolutional kernels to learn discriminative patterns that can be exploited to predict the failures of these machines. The raw information from the machines was preprocessed and converted into MTS. These were then used to train two models based on a feature learning approach to predict whether a failure would occur within the next seven days. The proposed predictive models employ convolutional kernels (MiniROCKET and HYDRA) to extract features from the original time series and a linear classifier to classify the sample based on the learned features. To evaluate the performance of these models, 30 executions of each classifier were run using different random seeds and data partitions. The average results of these experiments demonstrated that the MTS learned from the event-log information is correlated with the failure events, and thus, the novel dataset presented in this work is helpful in predicting future failure events. Most of the failures are challenging to predict, as the error patterns described in the event-log before a failure are complex and not easily detectable. Additionally, some failures are sudden and caused by misuse of the machine by its users. Despite these challenges, the experimental results achieved competitive performance and demonstrated how the proposed approach significantly overcame a state-of-the-art technique (i.e. InceptionTime) widely used for MTS classification and state-of-the-art ML non-temporal models commonly used in this context. The HYDRA model exhibited the best classification performance of the predictive models evaluated according to various metrics, including an AUC of 0.693. Additionally, statistical analysis revealed that the HYDRA, MiniROCKET and ROCKET models exhibited significant differences from the InceptionTime. Given the interest of the company in solving this problem, and taking into account the results of this study, the design of a DSS, that can be integrated into the existing company's infrastructure, has been proposed to assist the company in processing event-log data from the machines, generating the corresponding time series, and predicting when a failure is likely to occur. The DSS proposed in this study aims to automatically predict failure events, improving ATMs' performance and maintenance. To ensure the accuracy and reliability of the predictions, it is essential to include the expertise of the company's experienced technician, who should verify the machine for any potential issues in case an imminent failure event is detected. Integrating the proposed approach into a DSS led to developing a highly innovative, versatile tool that can be easily integrated into the company's infrastructure. The impact of this tool was also verified in collaboration with the Sigma company by confirming the following benefits:

- Positive feedback from the human expert operator for the output provided by the model on new unseen data.

- The DSS platform can communicate with the different devices distributed throughout the territory via the following communication protocols: REST and MQTT.

- The DSS platform, particularly the data processing module, is capable of continuous remote monitoring of the most indicative machine parameters while comput-ing the most discriminative pattern using the proposed convolution kernel-based approach.

- The development of a web-based and browser-usable graphical interface directly supports human operators.

- The DSS can be installed in the cloud or on proprietary data centers.

- The DSS can process a large amount of data by operating with variable bandwidths per device.

- The DSS can provide dedicated predictive maintenance reports.

Future work may be addressed to generalise the proposed approach to other domains (i.e. transports ticketing devices) while trying to mitigate even more consistently the naturalistic imbalanced setting of this task using ad-hoc resampling [61], cost-sensitive and ensemble strategies [62].

**Acknowledgement**

This work has been partially subsidized by "Agencia Española de Investigación (España)" (grant ref.: PID2020-115454GB-C22 / AEI / 10.13039 / 501100011033). Víctor Manuel Vargas's research has been subsidized by the FPU Predoctoral Program of the Spanish Ministry of Science, Innovation and Universities (MCIU), grant reference FPU18/00358. This work was supported by the research agreement between Università Politecnica delle Marche and Sigma S.p.A for the project "Smart Manufacturing Machine with Predictive Lifetime Electronic maintenance (SIMPLE)" funded by Ministero dello Sviluppo Economico (Italia) - Fondo per la Crescita Sostenibile - Accordi per l'innovazione di cui al D.M. 24 maggio 2017.

**Conflict of interest statement**

The authors declare that they have no competing interests.

## Appendix A. Results of the Wilcoxon tests

This appendix shows in Table A.7 the results of the Wilcoxon test for the other metrics which were not shown in Section 5.1. Note that the corrected $\alpha^* = 0.05/21 = 0.0024$. The $p$-value of the methodologies that show significant differences in performance are highlighted in bold font.

| Method | HYDRA | Inception | MR | RF | Ridge | ROCKET | XGB |
|---|---|---|---|---|---|---|---|
| | | | Accuracy | | | | |
| HYDRA | 1.000 | **0.002** | **0.001** | **< 0.001** | **< 0.001** | 0.171 | **< 0.001** |
| Inception | | 1.000 | 0.245 | **< 0.001** | **0.004** | **< 0.001** | **< 0.001** |
| MR | | | 1.000 | **< 0.001** | **< 0.001** | **< 0.001** | **< 0.001** |
| RF | | | | 1.000 | **< 0.001** | **< 0.001** | 0.688 |
| Ridge | | | | | 1.000 | **< 0.001** | **< 0.001** |
| ROCKET | | | | | | 1.000 | **< 0.001** |
| XGB | | | | | | | 1.000 |
| | | | Bal-Accuracy | | | | |
| HYDRA | 1.000 | **< 0.001** | **0.001** | **< 0.001** | **< 0.001** | **0.001** | **< 0.001** |
| Inception | | 1.000 | **< 0.001** | **0.019** | **< 0.001** | **< 0.001** | 0.082 |
| MR | | | 1.000 | **< 0.001** | **< 0.001** | 0.926 | **< 0.001** |
| RF | | | | 1.000 | **< 0.001** | **< 0.001** | 0.090 |
| Ridge | | | | | 1.000 | **< 0.001** | **< 0.001** |
| ROCKET | | | | | | 1.000 | **< 0.001** |
| XGB | | | | | | | 1.000 |
| | | | F1 | | | | |
| HYDRA | 1.000 | **< 0.001** | **0.003** | **< 0.001** | **< 0.001** | 0.090 | **< 0.001** |
| Inception | | 1.000 | **< 0.001** | **< 0.001** | **< 0.001** | **< 0.001** | **0.001** |
| MR | | | 1.000 | **< 0.001** | **< 0.001** | 0.052 | **< 0.001** |
| RF | | | | 1.000 | **< 0.001** | **< 0.001** | 0.262 |
| Ridge | | | | | 1.000 | **< 0.001** | **< 0.001** |
| ROCKET | | | | | | 1.000 | **< 0.001** |
| XGB | | | | | | | 1.000 |
| | | | MS | | | | |
| HYDRA | 1.000 | **< 0.001** | **0.037** | **< 0.001** | **< 0.001** | **< 0.001** | **< 0.001** |
| Inception | | 1.000 | **< 0.001** | **< 0.001** | **< 0.001** | **< 0.001** | **< 0.001** |
| MR | | | 1.000 | **< 0.001** | **< 0.001** | **0.001** | **< 0.001** |
| RF | | | | 1.000 | **< 0.001** | **< 0.001** | 0.491 |
| Ridge | | | | | 1.000 | 0.125 | **< 0.001** |
| ROCKET | | | | | | 1.000 | **< 0.001** |
| XGB | | | | | | | 1.000 |

Table A.7: Results of the Wilcoxon signed-rank test for the Accuracy, Bal-Accuracy, F1 and MS metrics. The $p$-values of the methods which are statistically different are marked with bold font. MR: MiniROCKET, IT: InceptionTime.